\pdfoutput=1
\documentclass[11pt]{article}

\usepackage[preprint]{acl}
\usepackage{times}
\usepackage{latexsym}

\usepackage[T1]{fontenc}
\usepackage[utf8]{inputenc}

\usepackage{microtype}

\usepackage{inconsolata}

\usepackage{graphicx}

\usepackage{algorithm}
\usepackage{amsmath}
\usepackage{booktabs}
\usepackage{multirow}
\usepackage{algpseudocode}

\DeclareMathOperator{\logprobsum}{logprobsum}

\title{Dynamic Depth Decoding: Faster Speculative Decoding for LLMs}

\author{Oscar Brown\textsuperscript{1,2}, Zhengjie Wang\textsuperscript{1}, Andrea Do\textsuperscript{1}, Nikhil Mathew\textsuperscript{1}, Cheng Yu\textsuperscript{1} \\
  \textsuperscript{1}ML Research Labs\thanks{This work was funded by Trellis Data Group. ML Research Labs is a subsidiary of Trellis Data Group.}, Canberra, Australia\\
  \textsuperscript{2}Australian National University
\\
  \small{
    \textbf{Correspondence:} \href{mailto:oscar.brown@mllabs.com.au}{oscar.brown@mllabs.com.au}, \href{mailto:zhengjie.wang@mllabs.com.au}{zhengjie.wang@mllabs.com.au}
  }
}

\begin{document}
\maketitle
\begin{abstract}
The acceleration of Large Language Models (LLMs) with speculative decoding provides a significant runtime improvement without any loss of accuracy. Currently, EAGLE-2 is the state-of-the-art speculative decoding method, improving on EAGLE with a dynamic draft tree. We introduce Dynamic Depth Decoding (DDD), which optimises EAGLE-2's tree drafting method using a dynamic depth. This extends the average speedup that EAGLE-2 achieves over EAGLE by $44\%$, giving DDD an average speedup of $3.16$x.
\end{abstract}

\begin{figure}[h]
    \centering
    \includegraphics[width=\linewidth]{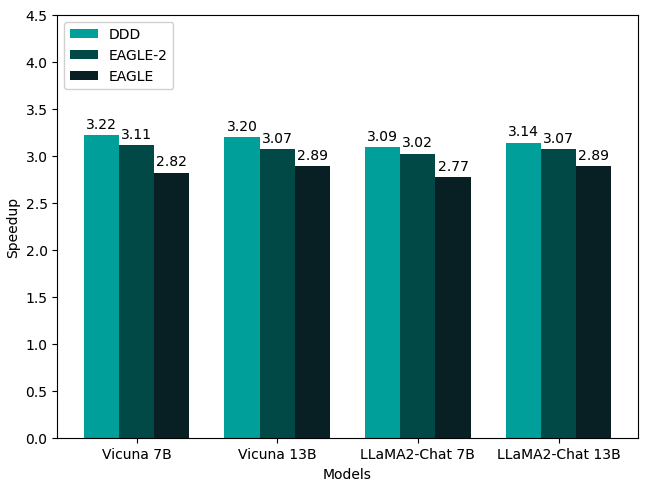}
    \caption{Speedup ratio compared to vanilla autoregressive decoding for different EAGLE decoding algorithms on MT-Bench with Temperature=0 on a single NVIDIA A40 GPU.}
    \label{fig:results}
\end{figure}

\section{Introduction}

Large Language Models (LLMs) \cite{brown2020language} \cite{touvron2023llama2openfoundation} have demonstrated impressive performance over various tasks. However, their large number of parameters causes inference speed to be too slow for many applications. 

Speculative Decoding \cite{leviathan2023fast} addresses this to accelerate an LLM, known as the target model. For each forward pass, the algorithm uses a much smaller draft model to generate a sequence of tokens to be inputted to the target model. Running the target model once is sufficient to verify the tokens until one is incorrect and generate the token that should follow the correct sequence. This gives a speedup by generating more tokens per forward pass of the target model. Notably, speculative decoding methods are lossless since every token is verified as correct by the target model.

\begin{figure}[h]
    \centering
    \includegraphics[width=\linewidth]{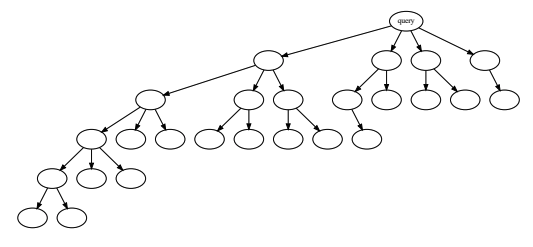}
    \caption{EAGLE's decoding tree \cite{li2024eagle}}
    \label{fig:decoding-tree}
\end{figure}

Extrapolation Algorithm for Greater Language-model Efficiency (EAGLE) \cite{li2024eagle} is the state of the art speculative decoding method, with it’s key feature being the construction of a draft model using the embedding layer and LM head of the target model with a single trainable head in between. On its first release, EAGLE used a method of generating a tree of tokens from the draft model and adjusting the target model’s attention mask to allow the entire tree to be inputted simultaneously into the target model. This tree has the structure shown in Figure \ref{fig:decoding-tree}, with the best tokens generated from each previous token being on the left. Although this tree chooses the tokens with the highest draft logprobs outputted after each token, its structure is static with no dependence on the draft model output.

EAGLE-2 \cite{li2024eagle2fasterinferencelanguage} improves on this static tree method by introducing a dynamic draft tree. The tree uses a beam search by choosing the top-k token sequences after each run of the draft model as the next input to the draft model. The sum of all logprobs generated in a sequence of tokens is used as a heuristic for choosing the top-k.

This approach is dynamic in that tokens are used, but it is not dynamic in the depth or width of the beam search. A dynamic width would not give a significant improvement because the runtime does not significantly increase as the beam search width increases (see Appendix~\ref{sec:appendix}). However, a dynamic depth would allow the draft model to be called a variable number of times per target model call, depending the likelihood of each sequence in the current beam being correct.

EAGLE-2's method is based on the finding that EAGLE's draft model confidence is a good approximation for the acceptance rate of draft tokens. This result is verified to also be true for other draft models \cite{liu2024optimizingspeculativedecodingserving}. Assuming this is true, we propose Dynamic Depth Decoding (DDD) to optimise EAGLE-2's decoding algorithm by making the beam search depth dynamic based on the draft confidence.

Numerous methods \cite{huang2024specdecboostingspeculativedecoding} \cite{mamou2024dynamicspeculationlookaheadaccelerates} \cite{liu2024parallelspeculativedecodingadaptive} have successfully used models to decide the draft decoding depth for standard speculative decoding, but none of these methods are designed for a tree-based algorithm such as EAGLE. We opt to use a heuristic based on the draft model confidence rather than a model because it is not evident that there is an effective way to input all the information from the tree into a model.

\section{Dynamic Depth Decoding}

DDD is an implementation of dynamic depth with EAGLE-2. We use the sum of the probabilities of all the sequences in the beam as a heuristic with a required threshold to continue draft generation. EAGLE-2 processes the draft model output with a logsoftmax function to produce logprobs for deciding the next beam. We can thus calculate the heuristic from the logprobs as in Equation \ref{equation:1} for a beam width \(w\) where the sum of logprobs of each sequence in the beam is $\logprobsum[i]$.

\begin{equation}
    \label{equation:1}
    H = \log(\sum_{i=0}^w \exp(\logprobsum[i]))
\end{equation}

While running the draft model, EAGLE-2 never uses data-dependent control flow, allowing the entire process to be lazy evaluated, providing significant optimisations. To determine whether to continue draft generation based on the probability sum, all the drafting must be immediately evaluated up to the current step. Hence, each time the heuristic is checked, there is a significant slowdown. To partially avoid this, we do not check the heuristic every step of the drafting process. Refer to Algorithm \ref{algorithm:dynamic-depth} for details of this method.

\begin{algorithm}[h]
    \caption{Dynamic Depth Decoding}
    \begin{algorithmic}[1]
            \Require a maximum number steps $n$, a beam width $w$, steps to check the heuristic $S$, a threshold to continue $x$
            \Ensure Postprocessed draft model output for any number of steps
            \State Run draft model on prompt
            \State Postprocess draft model output to produce beam using EAGLE-2's algorithm
            \State Set $\logprobsum$ to logprobs of beam
            \For {$\text{step} = 1$ to $n-1$}
                \If {$\text{step}$ in $S$}
                    \State $H = \log(\sum_{i=0}^w \exp(\logprobsum[i]))$
                    \If{$H<x$}
                        \State \textbf{break}
                    \EndIf
                \EndIf
                \State Run draft/ model on beam
                \State Postprocess draft model output to produce the new beam using EAGLE-2's algorithm
                \State Calculate $\logprobsum$ by adding the corresponding elements of $\logprobsum$ to logprobs of the new beam
            \EndFor
    \end{algorithmic}
    \label{algorithm:dynamic-depth}
\end{algorithm}

\section{Experiments}

\subsection{Setup}

We compare the speedup of DDD to both EAGLE and EAGLE-2 on a single NVIDIA A40 GPU. We test with temperature 0 on MT-bench. We use the same base and draft models as EAGLE-2 \cite{li2024eagle2fasterinferencelanguage} with Vicuna-7B, Vicuna-13B, LLaMA2-Chat 7B and LLaMA2-Chat 13B. We measure the speedup ratio against vanilla autoregressive decoding. We do not measure accuracy as every speedup method we test is lossless. We also do not measure token acceptance rate, since it is always better at a greater depth and does not test the effectiveness of a dynamic depth method.

\subsection{Parameters}

All parameters are kept constant across all our experiments. We use the optimal parameters for both EAGLE and EAGLE-2 from their papers \cite{li2024eagle} \cite{li2024eagle2fasterinferencelanguage}, including the draft tree shown in Figure \ref{fig:decoding-tree} for EAGLE and a depth of 6 for EAGLE-2. We have empirically found that DDD is optimal with a maximum of 11 draft model calls ($n=11$), with heuristic checks after the 5th, 7th and 9th steps ($S=\{5,7,9\}$) and a minimum logprob threshold of $x=-0.3$. Both EAGLE-2 and DDD are run with beam width $w=10$.

\subsection{Results}

On average over Table \ref{tab:results}, EAGLE-2 outperforms EAGLE by $8\%$, and DDD outperforms EAGLE-2 by $4\%$. However, EAGLE-2 is recorded to outperform EAGLE by $31\%$ in the same experiments \cite{li2024eagle2fasterinferencelanguage}. We therefore hypothesize that our hardware causes the gaps in speedup between decoding algorithms is reduced. We would encourage the developers of EAGLE to evaluate DDD on their hardware to provide a comparison under the same conditions that EAGLE and EAGLE-2 were evaluated. We note that the parameters of DDD may need to be optimised for their hardware.

\begin{table}[h]
  \centering
  \caption{Speedup ratios of different methods with Temperature=0 on MT-bench.}
  \begin{tabular}{ccc}
    \toprule
    Model & Method & Speedup\\
    \midrule
    \multirow{5}[2]{*}{Vicuna 13B} & EAGLE & 2.89x \\ & EAGLE-2 & 3.07x \\ & DDD & \textbf{3.20x} \\ & E-2 + Strict & 2.96x \\ & DDD + Strict & 3.07x \\
    \midrule
    \multirow{5}[2]{*}{LLaMA2-Chat 13B} & EAGLE & 2.89x \\ & EAGLE-2 & 3.07x \\ & DDD & \textbf{3.14x} \\ & E-2 + Strict & 2.95x \\ & DDD + Strict & 3.06x \\
    \midrule
    \multirow{5}[2]{*}{Vicuna 7B} & EAGLE & 2.82x \\ & EAGLE-2 & 3.11x \\ & DDD & \textbf{3.22x} \\ & E-2 + Strict & 2.83x \\ & DDD + Strict & 3.00x \\
    \midrule
    \multirow{5}[2]{*}{LLaMA2-Chat 7B} & EAGLE & 2.77x \\ & EAGLE-2 & 3.02x \\ & DDD & \textbf{3.09x} \\ & E-2 + Strict & 2.82x \\ & DDD + Strict & 2.92x \\
    \bottomrule
    \end{tabular}%
  \label{tab:results}%
\end{table}%

\subsection{Lazy Evaluation}

To determine the actual algorithmic advantage of DDD over EAGLE-2, we perform EAGLE-2's method as usual with depth $6$, but we break lazy evaluation between every call to the draft model by calling $\text{torch}.\text{cuda}.\text{synchronize}()$. We compare this to DDD with \(S=\{1,2,3,4,5,6,7,8,9,10\}\), where the heuristic is checked every step, also breaking lazy evaluation. We include this in Table \ref{tab:results} as "E-2 + Strict" and "DDD + Strict". We find that in this case, where lazy evaluation is broken every step, an average $5\%$ advantage can be achieved by DDD, across all the models. %This suggests that an algorithm that performs DDD without breaking lazy evaluation could theoretically achieve speedups of at least $1\%$ over DDD.

\section{Conclusion}
\label{sec:conclusion}

In this work, we introduce Dynamic Depth Decoding, an optimisation of EAGLE-2's decoding algorithm that increases the speedup of the current state-of-the-art speculative decoding method. We discover an opportunity to use the draft model's confidence to determine whether to continue drafting. Since the heuristic check breaks lazy evaluation, we find that it is optimal to check the heuristic only a few times. We also compare our decoding algorithm to EAGLE and EAGLE-2 over a variety of models. Future work on speculative decoding that significantly improves on the speedup of EAGLE-2 will most likely focus on optimising the draft model and the verification process, rather than the drafting algorithm.

\section{Limitations}
\label{sec:limitations}

We implement DDD with a series of breaks in lazy evaluation that causes a slowdown. Discovery of a way to perform an algorithm similar to DDD without the losses from breaking lazy evaluation would theoretically provide a significant advantage. Also, the results we observe on our hardware is significantly different to the published results of EAGLE-2 \cite{li2024eagle2fasterinferencelanguage}. Our model would be more easily compared with other methods if tested on their hardware.

\bibliography{main}

\begin{thebibliography}{9}
\providecommand{\natexlab}[1]{#1}

\bibitem[{Brown et~al.(2020)Brown, Mann, Ryder, Subbiah, Kaplan, Dhariwal, Neelakantan, Shyam, Sastry, Askell, Agarwal, Herbert-Voss, Krueger, Henighan, Child, Ramesh, Ziegler, Wu, Winter, Hesse, Chen, Sigler, Litwin, Gray, Chess, Clark, Berner, McCandlish, Radford, Sutskever, and Amodei}]{brown2020language}
Tom~B. Brown, Benjamin Mann, Nick Ryder, Melanie Subbiah, Jared Kaplan, Prafulla Dhariwal, Arvind Neelakantan, Pranav Shyam, Girish Sastry, Amanda Askell, Sandhini Agarwal, Ariel Herbert-Voss, Gretchen Krueger, Tom Henighan, Rewon Child, Aditya Ramesh, Daniel~M. Ziegler, Jeffrey Wu, Clemens Winter, Christopher Hesse, Mark Chen, Eric Sigler, Mateusz Litwin, Scott Gray, Benjamin Chess, Jack Clark, Christopher Berner, Sam McCandlish, Alec Radford, Ilya Sutskever, and Dario Amodei. 2020.
\newblock \href {https://arxiv.org/abs/2005.14165} {Language models are few-shot learners}.

\bibitem[{Huang et~al.(2024)Huang, Guo, and Wang}]{huang2024specdecboostingspeculativedecoding}
Kaixuan Huang, Xudong Guo, and Mengdi Wang. 2024.
\newblock \href {https://arxiv.org/abs/2405.19715} {Specdec++: Boosting speculative decoding via adaptive candidate lengths}.
\newblock \emph{Preprint}, arXiv:2405.19715.

\bibitem[{Leviathan et~al.(2023)Leviathan, Kalman, and Matias}]{leviathan2023fast}
Yaniv Leviathan, Matan Kalman, and Yossi Matias. 2023.
\newblock Fast inference from transformers via speculative decoding.
\newblock In \emph{International Conference on Machine Learning}, pages 19274--19286. PMLR.

\bibitem[{Li et~al.(2024{\natexlab{a}})Li, Wei, Zhang, and Zhang}]{li2024eagle2fasterinferencelanguage}
Yuhui Li, Fangyun Wei, Chao Zhang, and Hongyang Zhang. 2024{\natexlab{a}}.
\newblock \href {https://arxiv.org/abs/2406.16858} {Eagle-2: Faster inference of language models with dynamic draft trees}.
\newblock \emph{Preprint}, arXiv:2406.16858.

\bibitem[{Li et~al.(2024{\natexlab{b}})Li, Wei, Zhang, and Zhang}]{li2024eagle}
Yuhui Li, Fangyun Wei, Chao Zhang, and Hongyang Zhang. 2024{\natexlab{b}}.
\newblock Eagle: Speculative sampling requires rethinking feature uncertainty.
\newblock In \emph{International Conference on Machine Learning}.

\bibitem[{Liu et~al.(2024{\natexlab{a}})Liu, Li, Lv, Liu, Zhu, and Hu}]{liu2024parallelspeculativedecodingadaptive}
Tianyu Liu, Yun Li, Qitan Lv, Kai Liu, Jianchen Zhu, and Winston Hu. 2024{\natexlab{a}}.
\newblock \href {https://arxiv.org/abs/2408.11850} {Parallel speculative decoding with adaptive draft length}.
\newblock \emph{Preprint}, arXiv:2408.11850.

\bibitem[{Liu et~al.(2024{\natexlab{b}})Liu, Daniel, Hu, Kwon, Li, Mo, Cheung, Deng, Stoica, and Zhang}]{liu2024optimizingspeculativedecodingserving}
Xiaoxuan Liu, Cade Daniel, Langxiang Hu, Woosuk Kwon, Zhuohan Li, Xiangxi Mo, Alvin Cheung, Zhijie Deng, Ion Stoica, and Hao Zhang. 2024{\natexlab{b}}.
\newblock \href {https://arxiv.org/abs/2406.14066} {Optimizing speculative decoding for serving large language models using goodput}.
\newblock \emph{Preprint}, arXiv:2406.14066.

\bibitem[{Mamou et~al.(2024)Mamou, Pereg, Korat, Berchansky, Timor, Wasserblat, and Schwartz}]{mamou2024dynamicspeculationlookaheadaccelerates}
Jonathan Mamou, Oren Pereg, Daniel Korat, Moshe Berchansky, Nadav Timor, Moshe Wasserblat, and Roy Schwartz. 2024.
\newblock \href {https://arxiv.org/abs/2405.04304} {Dynamic speculation lookahead accelerates speculative decoding of large language models}.
\newblock \emph{Preprint}, arXiv:2405.04304.

\bibitem[{Touvron et~al.(2023)Touvron, Martin, Stone, Albert, Almahairi, Babaei, Bashlykov, Batra, Bhargava, Bhosale, Bikel, Blecher, Ferrer, Chen, Cucurull, Esiobu, Fernandes, Fu, Fu, Fuller, Gao, Goswami, Goyal, Hartshorn, Hosseini, Hou, Inan, Kardas, Kerkez, Khabsa, Kloumann, Korenev, Koura, Lachaux, Lavril, Lee, Liskovich, Lu, Mao, Martinet, Mihaylov, Mishra, Molybog, Nie, Poulton, Reizenstein, Rungta, Saladi, Schelten, Silva, Smith, Subramanian, Tan, Tang, Taylor, Williams, Kuan, Xu, Yan, Zarov, Zhang, Fan, Kambadur, Narang, Rodriguez, Stojnic, Edunov, and Scialom}]{touvron2023llama2openfoundation}
Hugo Touvron, Louis Martin, Kevin Stone, Peter Albert, Amjad Almahairi, Yasmine Babaei, Nikolay Bashlykov, Soumya Batra, Prajjwal Bhargava, Shruti Bhosale, Dan Bikel, Lukas Blecher, Cristian~Canton Ferrer, Moya Chen, Guillem Cucurull, David Esiobu, Jude Fernandes, Jeremy Fu, Wenyin Fu, Brian Fuller, Cynthia Gao, Vedanuj Goswami, Naman Goyal, Anthony Hartshorn, Saghar Hosseini, Rui Hou, Hakan Inan, Marcin Kardas, Viktor Kerkez, Madian Khabsa, Isabel Kloumann, Artem Korenev, Punit~Singh Koura, Marie-Anne Lachaux, Thibaut Lavril, Jenya Lee, Diana Liskovich, Yinghai Lu, Yuning Mao, Xavier Martinet, Todor Mihaylov, Pushkar Mishra, Igor Molybog, Yixin Nie, Andrew Poulton, Jeremy Reizenstein, Rashi Rungta, Kalyan Saladi, Alan Schelten, Ruan Silva, Eric~Michael Smith, Ranjan Subramanian, Xiaoqing~Ellen Tan, Binh Tang, Ross Taylor, Adina Williams, Jian~Xiang Kuan, Puxin Xu, Zheng Yan, Iliyan Zarov, Yuchen Zhang, Angela Fan, Melanie Kambadur, Sharan Narang, Aurelien Rodriguez, Robert Stojnic, Sergey Edunov, and Thomas
  Scialom. 2023.
\newblock \href {https://arxiv.org/abs/2307.09288} {Llama 2: Open foundation and fine-tuned chat models}.
\newblock \emph{Preprint}, arXiv:2307.09288.

\end{thebibliography}

\appendix

\section{Impact of Beam Width on Runtime}
\label{sec:appendix}

As the beam width of the target model increases, the runtime does not significantly change. Thus, an algorithm that dynamically changes the beam width would be unable to achieve a significant speedup over a constant beam width method such as EAGLE-2, which uses beam width 10.

\begin{table}[h]
  \centering
  \caption{Runtime per target model call for EAGLE-2 with depth 6 and Temperature=0 on MT-bench.}
  \begin{tabular}{ccc}
    \toprule
    Beam Width & Runtime (ms)\\
    \midrule
    5 & 51.36 \\
    10 & 51.41 \\
    20 & 51.42 \\
    50 & 52.39 \\
    75 & 53.91 \\
    100 & 54.33 \\
    \bottomrule
    \end{tabular}%
  \label{tab:width}%
\end{table}%

\end{document}